\RequirePackage{amsmath}
\documentclass[runningheads]{llncs}
\usepackage[T1]{fontenc}
\usepackage{microtype}
\usepackage{graphicx}
\usepackage{subcaption}
\usepackage{booktabs} 
\usepackage{hyperref}
\usepackage{orcidlink}
\RequirePackage{algorithm}
\RequirePackage{algorithmic}
\RequirePackage{fancyhdr}

\usepackage{amssymb}
\usepackage{mathtools}
\usepackage{amsthm}
\usepackage[capitalize,noabbrev]{cleveref}

\newcommand{\GAM}{$\textrm{GAM}_{\textrm{PINN}}$}
\newcommand{\MAML}{$\textrm{MAML}_{\textrm{PINN}}$}

\usepackage[textsize=tiny]{todonotes}
\begin{document}
\title{Meta-learning Loss Functions of Parametric Partial Differential Equations Using Physics-Informed Neural Networks}
%
\author{Michail Koumpanakis \,\orcidlink{0009-0008-2800-1693} \inst{1} \and
Ricardo Vilalta \,\orcidlink{0000-0001-8165-8805} \inst{2} }
%
\titlerunning{Meta-learning Loss Functions }
%
\institute{Department of Computer Science, University of Houston, Houston, TX, USA 
\and
Center for Science, Technology, Engineering, and Mathematics, University of Austin, Austin, TX, USA}
\maketitle              
\begin{abstract}
This paper proposes a new way to learn Physics-Informed Neural Network loss functions using Generalized Additive Models. We apply our method by meta-learning parametric partial differential equations, PDEs, on Burger's and 2D Heat Equations. The goal is to learn a new loss function for each parametric PDE using meta-learning. The derived loss function replaces the traditional data loss, allowing us to learn each parametric PDE more efficiently, improving the meta-learner's performance and convergence. 

\end{abstract}


\section{Introduction}
\label{introduction}

Neural Networks (NNs) have recently become widely accepted as an alternative way of solving partial differential equations (PDEs) due to their high efficiency in modeling non-linear and high-dimensional problems in a mesh-free data-driven approach. A prominent example is physics-informed neural networks (PINNs) \cite{RAISSI2019686}, where PDEs are effectively solved through a novel combination of data and domain knowledge. PINNs introduce a knowledge-informed loss that satisfies the application task's underlying physical laws.
\par
When solving multiple PDEs, a neural network must train a new model from scratch for every set of conditions. For example, Burger's equation \cite{Orlandi2000} is a PDE parametrized along different viscosities and initial and boundary conditions; a solution requires many computational steps.  A novel approach to address this problem is to use meta-learning \cite{Vanschoren2019} by teaching the model how to learn and generalize over a distribution of related tasks. Meta-learning methods divide the solution space into tasks, e.g., parametric PDEs. Learning a generalized representation of these tasks improves the convergence of the model on new tasks, e.g., fine-tuning with fewer iterations. Every task uses a few samples only, usually through one gradient-descent step, and provides feedback to the meta-learner. As the meta-learner improves, training on new tasks gives the model a head start, leading to faster convergence. 
\par
We suggest an additional step that teaches the model to learn the loss function at every task (i.e., meta-learns the loss function). Since loss functions play a significant role in the convergence of a neural network, this extension provides a different type of meta-knowledge to the meta-learner. Other approaches to choosing an appropriate loss function have been shown to improve a neural network's performance and convergence rate \cite{ciampiconi2023survey,lossfunctionsforDNNs}. 
\par
In this paper, we propose an approach that combines the popular meta-learning strategy of learning to initialize a model for fast adaptation with a new approach that meta-learns the loss function. The latter is attained by modeling the residuals of every meta-learning task using a Generalized Additive Model (GAM). We show that learning the loss function at every task improves the meta-learner's convergence on new tasks. Furthermore, we show that the GAM can be invoked to recover a loss function under noisy data. GAM's benefits can be ascribed to its role as a generalization term that improves model performance.

\section{Background and Related Work}

\subsection{Physics-Informed Neural Networks (PINNs)}

A common numeric approach to solve PDEs is to rely on Finite Element Methods (FEMs) \cite{FEM2014}. FEMs divide a high dimensional space into smaller, simpler units called finite elements, allowing the transformation of continuous problems into a system of algebraic equations by generating a finite element mesh. These equations are then solved iteratively to approximate the behavior of the physical system. Even though FEMs provide some versatility and can model complex boundary conditions, they have problems scaling to complex non-linear equations, requiring extensive computational power. A practical alternative is to use Physics-Informed Neural Networks (PINNs) by integrating physics-based constraints into the loss function. 
The network architecture is designed to satisfy a physical system's governing equations straightforwardly \cite{RAISSI2019686}. This ensures the solution adheres to the underlying physics while providing flexibility in handling complex, high-dimensional data.
 
PINNs have been applied to a plethora of scientific domains. Examples include inverse problems related to three-dimensional wake flows, supersonic flows, and biomedical flows \cite{Cai2021PhysicsinformedNN}; PINNs can be used as an alternative method to solve ill-posed problems, e.g., problems with missing initial or boundary conditions.
One example is that of heat-transfer problems \cite{Cai2021,Zhili2021}, where PINNs show remarkable performance over traditional approaches when solving real-industry problems efficiently with sparse data. PINNs can also be applied to astrophysical tasks; in one study, PINNs are used to model astrophysical shocks with limited data~\cite{Moschou_2023}; the study shows model limitations and suggests a data normalization method to improve the model's convergence.

\subsection{Meta-Learning}

Meta-learning has emerged as a critical technique in machine learning that enables knowledge transfer across tasks with low data requirements. A popular approach optimizes the initialization of model parameters to facilitate quick adaptation through a small number of gradient steps at every new task~\cite{maml}. This meta-learning strategy has been successfully applied to multiple domains, such as reinforcement learning \cite{Nagabandi2018LearningTA}, computer vision \cite{ren2018meta}, and natural language processing \cite{gu-etal-2018-meta}. \par
In solving parametric partial differential equations (PDEs),
previous work \cite{PENWARDEN2023} proposes a new meta-learning method for PINNs that computes initial weights for different parametrizations using the centroid of the feature space~\cite{PENWARDEN2023}. Another line of work proposes different neural network architectures for meta-learning of parametric PDEs \cite{MAD,gpt_pinn}. For example, one can use a meta-auto-encoder model to capture heterogeneous PDE parameters as latent vectors; the model can then learn an approximation based on task similarity~\cite{MAD}. Another work uses a generative neural network (GPT-PINN) with customized activation functions in the hidden layer that act as pre-trained PINNs instantiated by parameter values chosen by a greedy algorithm~\cite{gpt_pinn}.
An interesting approach to solving parametric PDEs is to meta-learn the PINN loss function~\cite{meta_learn_pinn_loss}. The idea is to encode information specific to the considered PDE task distribution while enforcing desirable properties on the meta-learned model through novel regularization methods. 

Unlike previous work, our approach focuses on modeling the residuals of every task using a GAM to attain accurate models resilient to noisy data. The proposed approach centers on learning the specific data-loss function and improving the meta-learner's adaptability and generalization to new tasks. 

\subsection{Generalized Additive Models (GAMs)}

Generalized Additive Models (GAMs) \cite{GAMs} have gained popularity among regression techniques for their ability to model complex relationships and generate flexible data representations. GAMs allow additive combinations of smooth functions to capture linear and non-linear dependencies.
A GAM can be defined as:
\begin{equation}
f(x) = f_1(v_1) + f_2(v_2) + \ldots + f_n(v_n) \label{eq:gam_definition}
\end{equation}
where each $f_i$ describes a smooth function that maps the $i$-th input feature, $v_i$ (or combination of features), to the output space \cite{KRAUS2023}.
An empirical evaluation of the predictive qualities of numerous GAMs compared to traditional machine learning models assesses model performance and interpretability \cite{GAMchanger}. The study shows how advanced GAM models such as EBM \cite{EBM} or GAMI-Net \cite{GAMI-Net} often outperform traditional white-box models, e.g., decision trees, and perform similarly to conventional black-box models, e.g., neural networks.

\section{Methodology}

Our methodology solves PDEs (e.g., the viscous Burgers equation and the 2D Heat equation) using Physics-Informed Neural Networks (PINNs), Meta-Learning for fine-tuning new PDEs, and Generalized Additive Models (GAMs) for learning loss functions across tasks. PINNs are used to solve each parametric PDE, and the meta-learning algorithm is used to learn diverse representations across tasks, leading to faster convergence. 

We focus on the data loss; we use GAMs to generate a new loss term by learning the model residuals for each parametric PDE. The goal is to efficiently handle initial and boundary conditions and accelerate the training of new partial differential equations (PDEs). Due to their additive nature, GAMs can facilitate the discovery of additional terms in the loss function. Next, we detail the different modules of our proposed architecture. 

\subsection{Fast-Model Adaptation}

We employ Model-Agnostic Meta-Learning~\cite{maml}, MAML, to efficiently initialize the neural network weights for different parametric PDEs. The meta-learning process involves offline optimization of the network weights with a few examples from other tasks as a learning process. Specifically, the meta-objective function for MAML is defined as the sum of losses over multiple tasks:
\begin{equation}
    \min_{\theta} \sum_{\text{tasks}} \mathcal{L}(\hat{u}(\theta)),
    \label{eq:maml_objective}
\end{equation}
where $\hat{u}(\theta)$ denotes the solution of the PDE, and $\mathcal{L}$ is the loss function. 

The meta-learner is trained on tasks during the meta-training stage, each with training and testing data (support and query sets). The support set is used to train the model on the current task. It consists of a few labeled examples (few-shot learning), e.g., 5 or 10 examples. The query set evaluates the model’s performance on the current task after it has been trained on the support set; it simulates the model’s ability to generalize to new, unseen examples within the same task.
It contains examples that are not part of the support set but of the same task distribution. These examples compute the loss and update the meta-learner during training.

We perform task-specific meta-training for each parametric task defined by a unique set of parameters $\theta$. In the meta-testing stage, we initialize the network weights of every new test task using the pre-trained MAML weights and fine-tune the model with task-specific data.

\subsection{Incorporating Physical Constraints}

Traditional neural networks rely on a single residual loss $L(\hat{u}, u)$, usually computed as the mean squared error (i.e., L2 loss). PINNs define an additional physics-informed loss that minimizes the residuals of the PDE, i.e., $L(E(\hat{u}))$ where E is the function to be minimized. If we consider an arbitrary PDE dependent on $u(x,t;w)$:

\begin{equation}
     Z(u, \nabla u,\nabla^2 u,..,\nabla^n u) = F(u)  
    \label{eq:pde_form}
\end{equation}

\noindent
then its residual error $E$ is defined as:

\begin{equation}
    E = Z(u, \nabla u,\nabla^2 u,..,\nabla^n u) - F(u)
    \label{eq:residual_form}
\end{equation}

\noindent The total loss function of a physics-informed neural network is then given by:

\begin{equation}
    \mathcal{L}_{\text{total}} = \mathcal{L}_{\text{PDE}} + \mathcal{L}_{\text{data}},
    \label{eq:loss}
\end{equation}
\begin{equation}
\mathcal{L}_{\text{data}} = \text{MSE}_{ib} = \frac{1}{N_{ib}} \sum_{i=1}^{N_{ib}} \left( \hat{u}(x_i^{ib}, t_i^{ib}) - u_i \right)^2
\label{eq:loss_data}
\end{equation}
\begin{equation}
\mathcal{L}_{\text{PDE}} = \text{MSE}_R = \frac{1}{N_R} \sum_{i=1}^{N_r}E( \hat{u}(x_i^r, t_i^r))^2
\label{eq:loss_pde}
\end{equation}

\noindent
where ${x_i^{ib}, t_i^{ib}}$ are the initial and boundary condition points and ${x_i^{r}, t_i^{r}}$ the inner collocation points. The set of ${u_i, \hat{u_i}}$ are the -initial and boundary conditions- ground truth and predicted velocity values. The data are randomly defined in a mesh-free way as a collection of inner collocation points $N_r$ and initial and boundary points $N_{ib}$. 
$\mathcal{L}_{\text{PDE}}$ is the physics-informed loss while $\mathcal{L}_{\text{data}}$ is the data loss. For the inner collocation points, the solution $u(x,t)$ is unknown; the residual physics-informed loss is used to minimize the error. We use common optimization techniques for neural networks such as ADAM and L-BFGS.

\subsection{Proposed Architecture}

We propose using GAMs to estimate the model residuals, providing an additional layer of flexibility and expressiveness to the model. This can be characterized as sequential residual regression (SRR)~\cite{SRR}.
Sequential residual regression builds a model step-by-step; each step focuses on fitting a simple model (a single feature or a small subset of features) to the residuals left by the current model. The idea is to iteratively improve the model by addressing the remaining unexplained variance (residuals) in the data. These models have been shown to improve model interpretability and manage high-dimensional data settings~\cite{sequentialmodelopt}.  
They share conceptual similarities with boosting~\cite{Boosting}, where model residuals are used as input to build the next model until a strong learner is attained.  
The proposed approach reduces over-fitting and brings a regularization term to the loss function. 

The GAM provides a semi-symbolic expression, i.e., additional terms in the loss function, that we use to replace the traditional data loss (MSE) for every MAML task. This involves capturing the loss of the initial and boundary conditions as a function of the input features. This approach enables the model to estimate the average loss more accurately, improving overall predictive performance.

We compute the residuals as $err = \hat{u} - u$ for the initial and boundary data. We fit the boundary and initial condition input features (spatial and temporal) against the residuals using a GAM model (see Eq.~1). The predicted loss, averaged over all the data points, $\mathcal{L}_{\text{GAM}}$, is then added to the PDE loss:

\begin{equation}
    \mathcal{L}_{\text{total}} = \mathcal{L}_{\text{PDE}} + \mathcal{L}_{\text{GAM}},
    \label{eq:total_loss_gam}
\end{equation}

Algorithm \ref{alg:algorithm_pseudocode} provides a detailed explanation of our approach. We split our data into a meta-training set $S_{tr}$ and a meta-testing set $S_{te}$. We sample tasks from the meta-training set to train our meta-learner. A separate neural network is trained for every task to predict the solution for the inner collocation points and the initial and boundary data.
The support data are used to train the model, and the query data are used to evaluate it. The GAM loss is then computed from the initial and boundary data residuals, which acts as an additive loss function for the support loss. i.e., the GAM loss is added to the PDE loss, after which a gradient-descent step is invoked to optimize the network. 

In the task evaluation stage, the query set is
used to evaluate the model’s performance on the current task. The average query loss is computed for every task, i.e., $L_{meta}$, which is used to optimize the meta-learner. We compute the GAM loss only for the support/train data of the meta-learner's tasks. This allows us to measure the performance of the two different meta-learning approaches on equal terms at the query stage.
Finally, in the meta-testing stage, we sample tasks from the meta-testing set $S_{te}$ and evaluate the two meta-learning approaches. We compute the mean squared loss (MSE) overall meta-testing tasks and compare our results.

\begin{algorithm}[tb]
   \caption{Meta-learning with PINNs}
   \label{alg:algorithm_pseudocode}
\small 
\begin{algorithmic}[1]
   \STATE {\bfseries Require:} meta-training set $S_{tr}$, meta-testing set $S_{te}$
   \STATE {\bfseries Require:} convergence criteria $\epsilon = 10^{-3}$
   \STATE {\bfseries Require:} meta-learner's initial loss $L_{meta} > \epsilon$
   \STATE {\bfseries Require:} number of training iterations N
   \STATE Randomly initialize $\theta$
   \STATE Sample n tasks $T \sim S_{tr}$
   \WHILE{$L_{meta} > \epsilon$ and counter < N}
   \FOR{each task $T_i$}
   \STATE Sample support $D^{S}_{T_i}$ and query $D^{Q}_{T_i}$ sets from $T_i$ with k data points each
   \STATE Predict $\hat{u} = NN (D^{S}_{T_i},\theta)$
   \STATE $L_{pde} = MSE(E(\hat{u}))$, defined in (4)
   \STATE Compute residuals $r = \hat{u} - u$
   \STATE Compute $F_{gam_i} = GAM (r, D^{S}_{T_i})$
   \STATE $L_{gam} =$ average of applying ${F}_{gam_i}$ over the $k$ data points
   \STATE Compute support loss $L^{\text{S}}_{T_i} = L_{pde} + L_{gam}$
   \STATE Perform inner-loop optimization using support data \small $D^{S}_{T_i}$:
   \vspace{-0.5\baselineskip}
    \[
     (\theta'_{i}, W'_{i}, b'_{i}) \leftarrow (\theta_{i}, W_{i}, b_{i}) - \alpha \nabla(\theta_{i}, W_{i}, b_{i}) L^{\text{S}}_{T_i}
    \]
    \STATE Evaluate generalization performance on query data $D^{Q}_{T_i}$:
    \STATE $L_{data} = MSE(\hat{u}, u)$
    \[
    L^{Q}_{T_i} = L_{pde} + L_{data} 
    \]
    \ENDFOR
    \STATE Compute meta-learner loss on n tasks:
   $\mathcal{L}_{meta} = \frac{1}{n} \sum_{D_{T_i}} \mathcal{L}^{Q}_{T_i}$
   \STATE Compute gradient $\nabla_{(\theta, W, b)} \mathcal{L}_{meta}$ across batch of meta-training tasks
    \STATE Perform outer-loop optimization to update meta-learner
    \ENDWHILE
    \STATE For meta-testing stage, sample m tasks $T \sim S_{te}$
    \FOR{each task $T_j$}
    \STATE Fine-tune network as initialized by the meta-learner and compute $\mathcal{L}_{T_j}$
    \STATE Compute average loss $\mathcal{L}_{test} = \frac{1}{m} \sum_{D_{T_j}} \mathcal{L}_{T_j}$
    \ENDFOR
\end{algorithmic}
\end{algorithm}
\vspace{-10pt}

\section{Experiments}

This section shows the results of applying different learning models to two families of equations, the 1D Viscous Burgers equation and the 2D Heat equation, with parametric initial conditions. The model's accuracy is measured using the Mean Squared Error (MSE). We provide the mean value of the MSE when fine-tuning new tasks.  
\footnote{Our code is available at: \url{https://github.com/Mkoumpan/GamPINN}}

Each experiment's PDE parameters are divided into sample tasks for meta pre-training $S_{tr}$ and new sample tasks for fine-tuning $S_{te}$. The methods under comparison are the following:

\begin{itemize}
  \item \textbf{Random-Weighting:} Trains the model with random weights from scratch based on the PINNs method for all PDE parameters in $S_{te}$, task-by-task.
  \vspace*{2mm}
  \item \textbf{\MAML:} Meta-trains the model for all PDE parameters in $S_{tr}$ based on the MAML algorithm. In the meta-testing stage, load the pre-trained weights \(\theta^*\) and fine-tune the model for each PDE parameter in $S_{te}$.
  \vspace*{2mm}
  \item \textbf{\GAM:} Our proposed approach; it meta-trains the model for all PDE parameters in $ S_{tr}$ using the MAML algorithm. For every task/PDE parameter, we fit the residuals of the initial and boundary conditions to a GAM model and derive a new data loss. We use the GAM model to provide an additional term for the PINN loss.

\end{itemize}

\subsection{1D Viscous Burgers equation}

The viscous Burgers equation is given by
\begin{equation}
    \frac{\partial u}{\partial t} + u \frac{\partial u}{\partial x} = \nu \frac{\partial^2 u}{\partial x^2},\quad x \in X,\, \ t \in T
    \label{eq:burgers}
\end{equation}
\[
u(x, 0) = u_{0}(x),\quad x \in X
\]
where $X \in [-1, 1]$ and $T \in [0, 1]$. The term $u(x, t)$ represents the fluid velocity, and $\nu = 0.05$ is the viscosity coefficient, which is constant in our case. The boundary conditions at points $x=1$ and $x=-1$ are equal to 0. $u(1, t) = u(-1, t) = 0$. We consider variable initial conditions of the form:
\begin{equation}
    u(x, 0; \theta) = - \sin(\pi x) + \theta \cos(\pi x),
    \label{eq:parametric_ic}
\end{equation}
where $\theta$ introduces variability/parametrization, and different values of $\theta$ define distinct MAML tasks and distinct PDE solutions. The parameter $\theta$ is sampled from a uniform distribution $p(\theta) = U(0, 1)$. We train our meta-learner using five tasks with only one gradient descent step for each epoch, equivalent to a 5-shot 1-way meta-learning problem. 

We use a neural network architecture with seven hidden layers, 20 hidden nodes at each layer, and an Adam optimizer with a learning rate of $0.005$. When training from scratch, we use randomly distributed points {$N_f$ = 10000, $N_{ib}$ = 100}, where $N_f$ are the inner collocation points and $N_{ib}$ the initial and boundary condition points. For \MAML\ and \GAM\, we use random {$N_f = 20$, $N_{ib}$ = 10} for the support set and the same number of points for the query set; we train the neural network for 7,000 epochs. Using 2,000 epochs, we evaluate our results on 10 new tasks sampled from $p(\theta)$.

\begin{table}[ht!]
\caption{MSE performance at epoch 1000, 1500, and 2000 of different tasks using the explained meta-learning techniques. The last column compares \GAM\ with RANDOM and \MAML\, respectively; each asterisk shows a statistically significant difference at the p-value of 0.02 using a one-tailed t-student test.}
\label{sample-table}
\vskip 0.01in
\begin{center}
\resizebox{\linewidth}{2.7cm}{%
\begin{tabular}{lccccr}
\toprule
Method & Epoch($10^3$) & \multicolumn{4}{c}{MSE} \\
\cmidrule(lr){3-6}
& &$Task_1$ $(\theta = 0)$ & $Task_2 $ $(\theta = 0.3)$ & $Task_3$ $(\theta = 0.7)$ & $Task_{mean}$\\
\midrule
RANDOM    & 1 & 0.018 & 0.027 & 0.05 & 0.039\\
\MAML\ & 1& 0.014 & 0.015 & 0.033 & 0.028\\ 
\GAM\ & 1 &0.002 & 0.0028 & 0.0093 & $0.0079^{**}$\\
\midrule
RANDOM    & 1.5 & 0.014 & 0.022 & 0.043 & 0.033\\
\MAML\ & 1.5 & 0.0022 & 0.0033 & 0.011 & 0.0083\\ 
\GAM\ & 1.5 &0.009 & 0.0017 & 0.007 & $0.0059^{**}$\\
\midrule
RANDOM    & 2 & 0.002 & 0.02 & 0.035 & 0.017\\
\MAML\ & 2& 0.001 & 0.002 & 0.009 & 0.0066\\ 
\GAM\ & 2 &0.0006 & 0.001 & 0.006 & $0.0047^{**}$\\
\bottomrule
\end{tabular}
}
\end{center}
\vskip -0.1in
\end{table}

\begin{figure}[ht!]
\vskip 0.2in
\begin{center}
\centerline{\includegraphics[width=0.8\columnwidth]{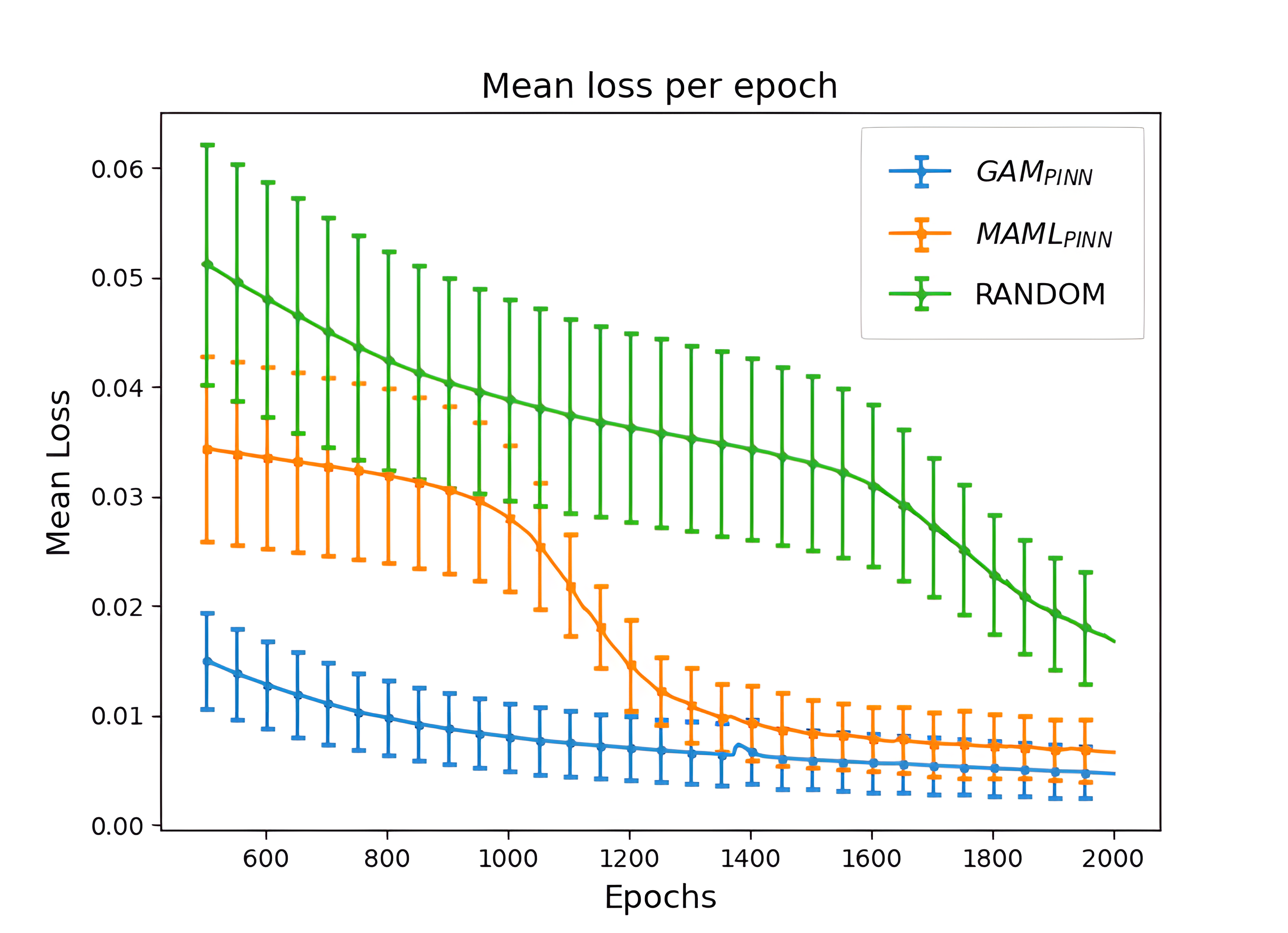}}
\caption{Burgers’ equation: The convergence of mean loss overall parametric values for the number of training iterations. Confidence intervals (95\%) assuming a normal distribution indicated by error bars illustrate the uncertainty in the loss measurements.}
\label{icml-historical}
\end{center}
\vskip -0.2in
\end{figure}

\begin{figure}[ht!]
\vskip 0.2in
\begin{center}
\centerline{\includegraphics[width=0.9\columnwidth]{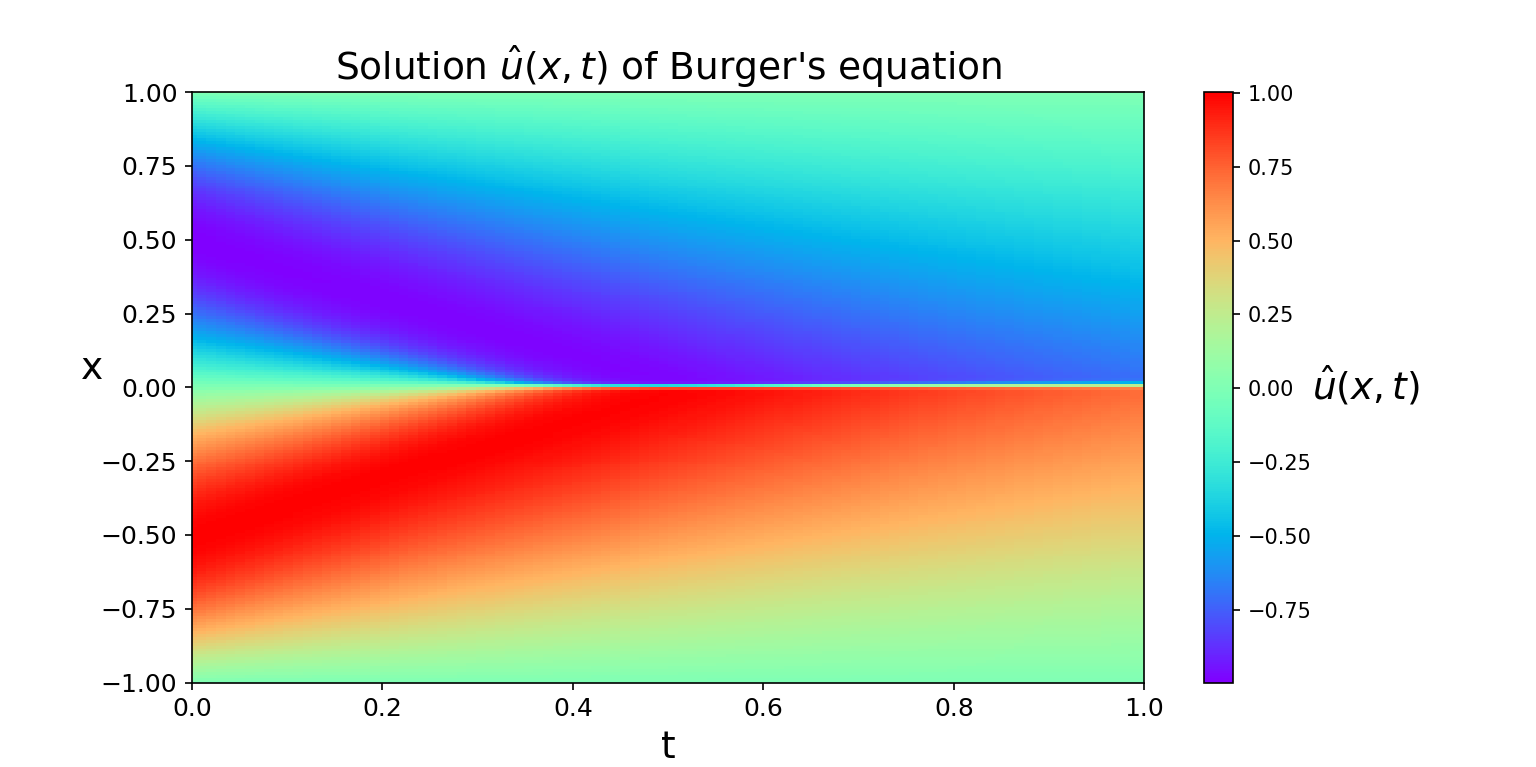}}
\caption{Burgers’ equation: Solution of $\hat{u}_{xt}$ with periodic initial conditions ($\theta$ = 0). The solution approximates the ground truth of the equation with an error less than $5e^{-4}$.}
\end{center}
\vskip -0.2in
\end{figure}

Table 1 shows how \GAM\ outperforms Random Weighting, taking fewer epochs to converge to a solution than \MAML.\ 
Figure~1 shows the MSE of all methods as the number of training iterations increases (fine-tuning stage). All methods eventually converge to nearly the same accuracy (the MSE being close to 0.001). Regarding convergence, Random Weighting needs more than 2000 iterations, whereas \MAML\ needs more than 1500, and \GAM\ needs about 1100 iterations. \GAM\ performs best across all tasks, converging fast. 

Figure~2 shows the solution of Burger's equation with a parameter $\theta = 0$ on the initial condition (Eq.~11). The density plot describes how the fluid velocity $u(x, t)$ changes regarding the position $x$ inside the 1-D tube and the moment in time $t$ it's measured. The color bar on the right describes the different values of the velocity, i.e., blue (-1) < green (0) < red (1).

\subsection{2D Heat Equation}
The 2D heat equation is given by
\begin{equation}
    \frac{\partial^2 u}{\partial x^2} + \frac{\partial^2 u}{\partial y^2} = \frac{\partial u}{\partial t}, \quad x \in X, \ y \in Y, \ t \in T
    \label{eq:2d_heat}
\end{equation}
\[
u(x, y, 0) = u_{0}(x, y),\quad x \in X, \ y \in Y
\]
where $X \in [-1, 1]$, $Y \in [-1, 1]$, $T \in [0,1]$. 
The term $u(x, y, t)$ represents the temperature inside the 2D plate. Dirichlet boundary conditions are used for the PDE, where:
\begin{equation}
    u(x,y=1,t) = \sin(\pi x),
\end{equation}
and the three other edges are equal to 0.
This represents a cold plate periodically heated through its top edge on the x-axis.
We run experiments under two different periodical initial conditions, parameterized amplitude $a_n$ (Eq.~\ref{eq:parametric_ic_heat_1}) and parameterized frequency $b_n$ (Eq.~\ref{eq:parametric_ic_heat_2}). We consider variable initial conditions of the form:

\begin{equation}
    u(x, y, 0; a) = a_{1}\sin(\pi x) + a_{2}\cos(\pi x),
    \label{eq:parametric_ic_heat_1}
\end{equation}

\begin{equation}
    u(x, y, 0; b) = \sin(b_1\pi x)*\cos(b_2\pi x),
    \label{eq:parametric_ic_heat_2}
\end{equation}

Different values of a and b describe variability/parametrization of the initial condition and, in turn, distinct MAML tasks and PDE solutions. We choose different initial conditions than the previous experiments, i.e., two parametrizations, to increase the complexity of the PDE. The parameters $a$ and $b$ are sampled from a uniform distribution $p(a,b)$ = U(0, 1).
The training of the meta-learner and the model's hyperparameters are the same as in the previous experiment.

\begin{figure}[ht!]
\begin{center}
\centerline{\includegraphics[width=0.8\columnwidth]{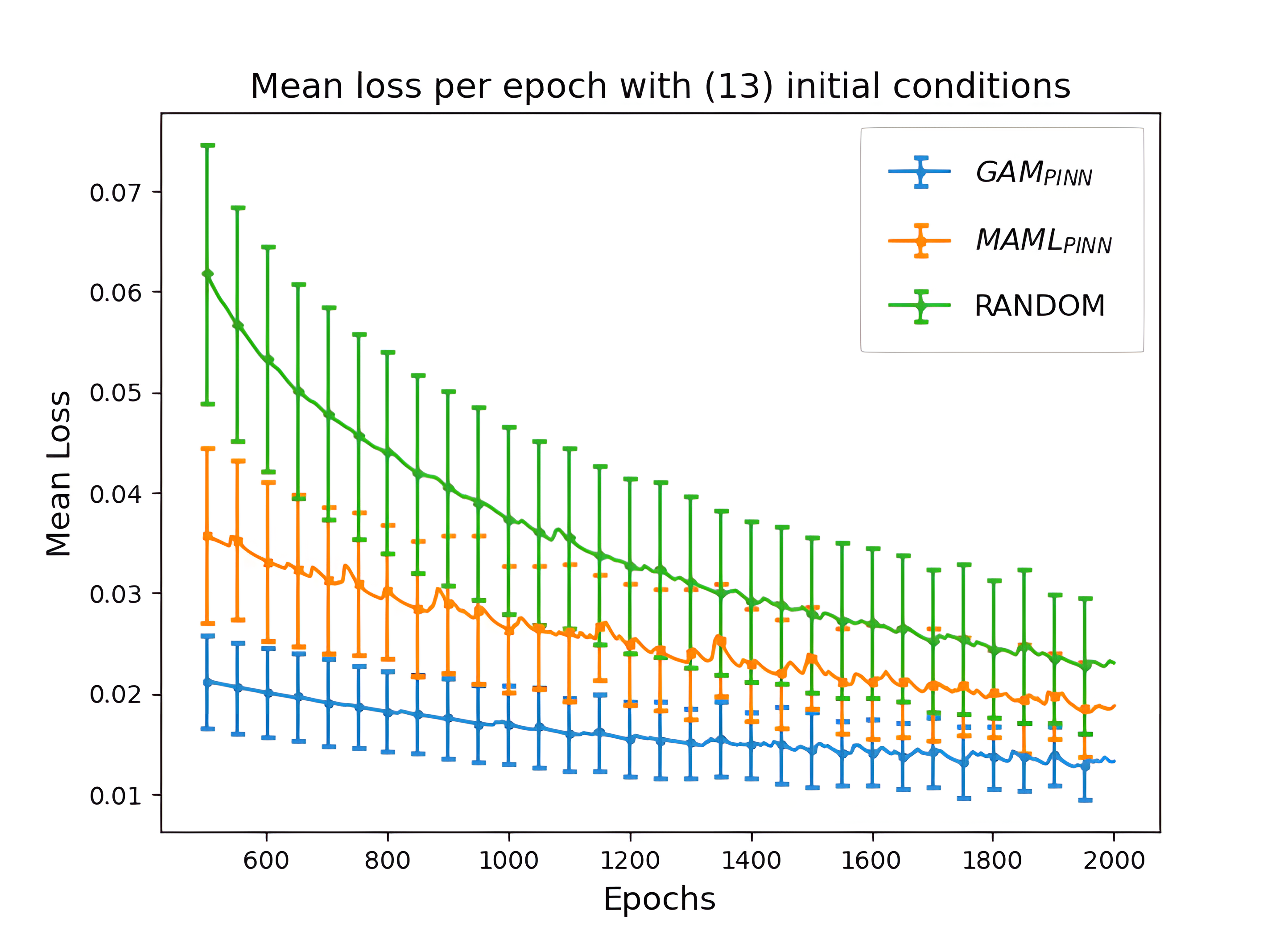}}
\caption{2D Heat equation: The convergence of mean loss overall parametric values concerning the number of training iterations using initial conditions defined in Eq.~13.}
\label{2d_heat_loss_default}
\end{center}
\vskip -0.2in
\end{figure}

In Figures~3 and~4, we can see the convergence of the mean loss over the number of training iterations for two different initial conditions. Like Burger's equation, we see how \GAM\ outperforms \MAML\ and Random Weight initialization with a $\sim 50\%$ increase in performance. Specifically, using initial conditions (Eq.~\ref{eq:parametric_ic_heat_2}), Figure~4 shows how $\textrm{MAML}_{\textrm{PINN}}$ cannot converge faster than Random Weight initialization. Using a GAM as a residual modeler makes the convergence of all tasks significantly faster.

\begin{figure}[ht!]
\begin{center}
\centerline{\includegraphics[width=0.8\columnwidth]{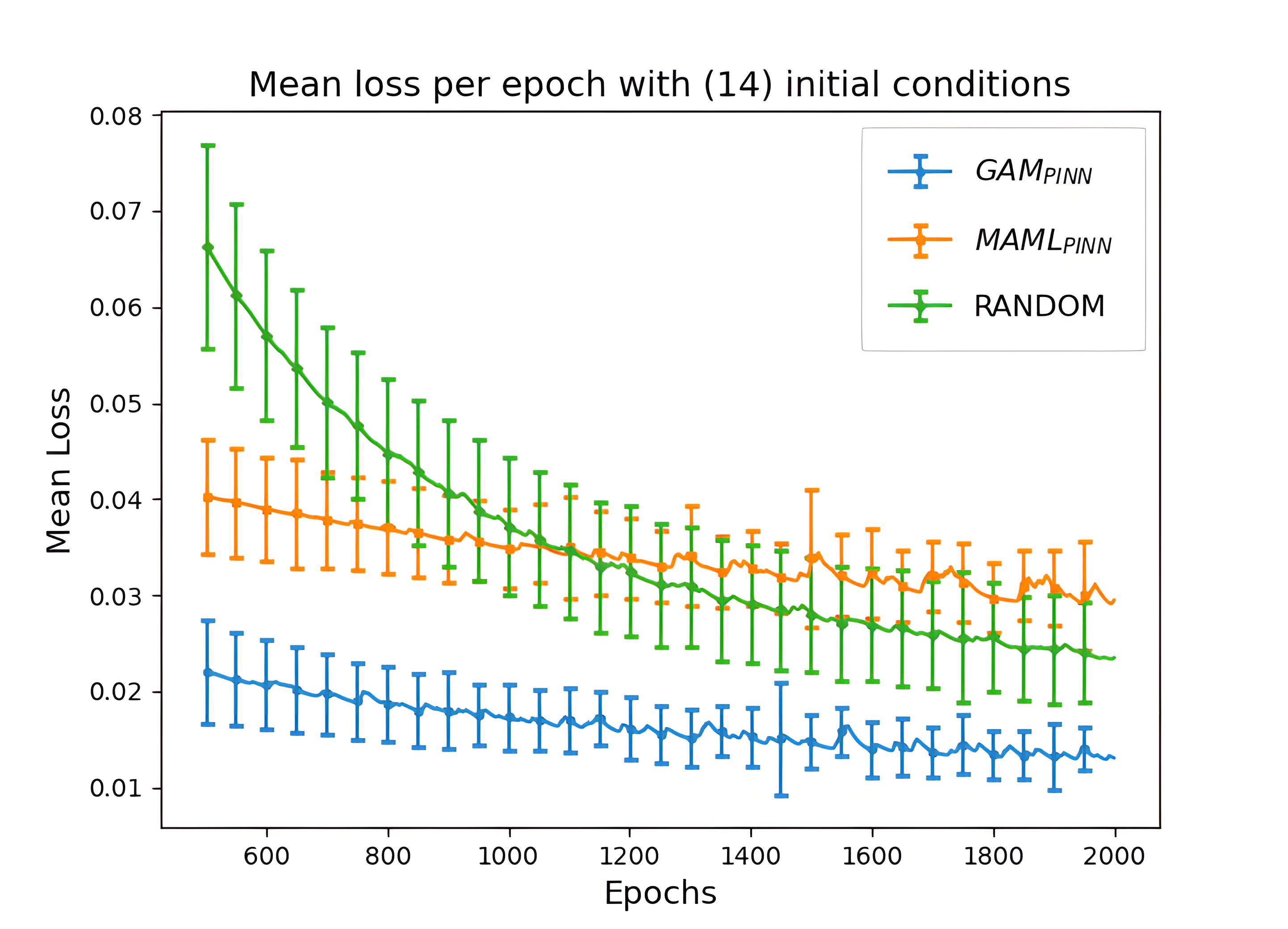}}
\caption{2D Heat equation: The convergence of mean loss overall parametric values concerning the number of training iterations using initial conditions defined in Eq.~14. Confidence intervals (95\%) assuming a normal distribution that are indicated by error bars illustrate the uncertainty in the loss measurements.}
\label{2d_heat_loss_multiply_init}
\end{center}
\vskip -0.2in
\end{figure}

\subsection{Handling Noise in Burgers' Equation}

\begin{figure}[ht!]
\centering
\begin{subfigure}{\linewidth}
    \centering
    \includegraphics[width=0.8\columnwidth]{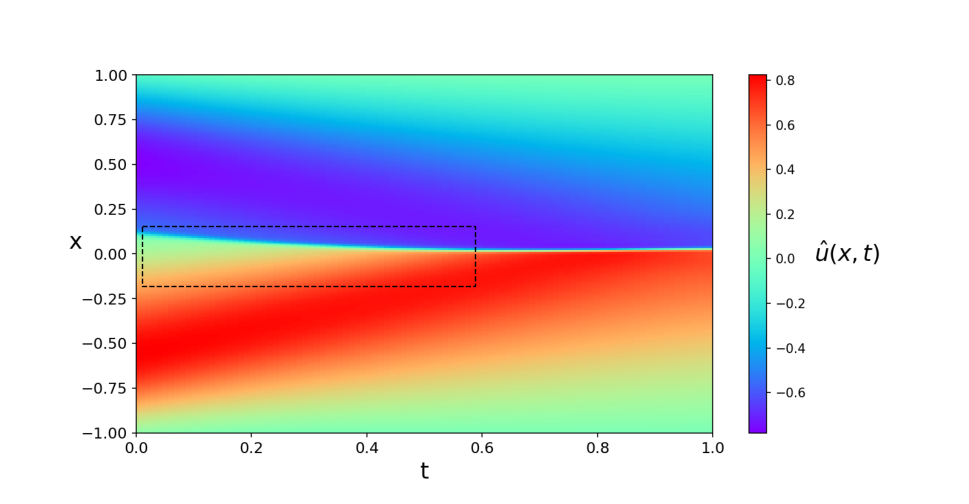}
    \label{fig:subfigA}
\end{subfigure}
\vskip -0.1in
\begin{subfigure}{\linewidth}
    \centering
    \includegraphics[width=0.8\columnwidth]{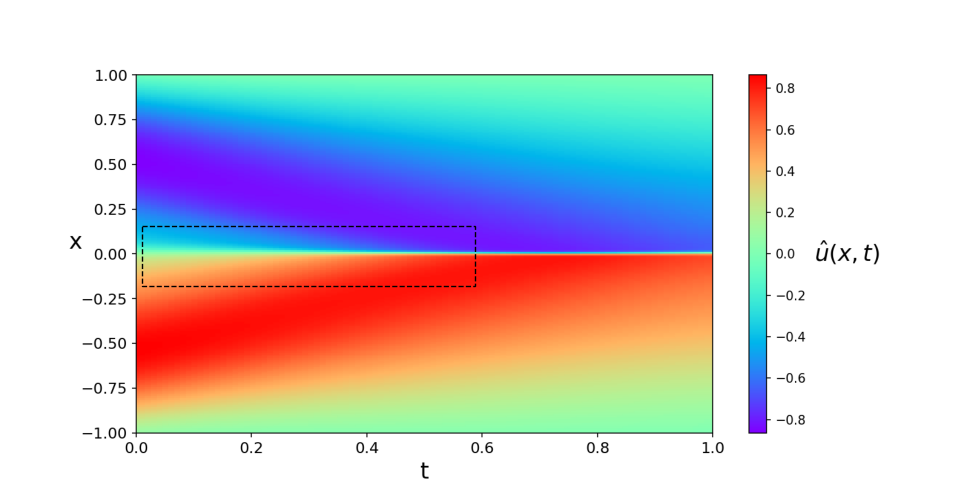}
    \label{fig:subfigB}
\end{subfigure}
\caption{Burgers' equation: comparison of the noisy solution u(x,t) (top) with de-noised solution (bottom) for initial conditions (Eq.~15) with noise $p = 5\%$.}
\label{fig:subfigures}
\label{noisy_burgers_0.05}
\end{figure}

We further show the benefits of the GAM function. We invoke the noisy viscous Burgers' equation defined as
\begin{equation}
    \frac{\partial u}{\partial t} + u \frac{\partial u}{\partial x} = \nu \frac{\partial^2 u}{\partial x^2} + p\epsilon ,\quad x \in X,\, t \in T
    \label{eq:burgers_noise_2}
\end{equation}
\[
u(x, 0) = u_{0}(x),\quad x \in X
\]
where $X \in [-1, 1]$,\ $T \in [0, 1]$,\ $\epsilon \in [-1,1]$. $u(x, t)$ represents the fluid velocity, $\epsilon$ is the random noise, p is a user-defined hyper-parameter that weights the noise contribution, and  $\nu = 0.05$ is the viscosity coefficient, constant in our case. The boundary conditions at points $x = 1$ and $x = -1$ are equal to 0. $u(1, t) = u(-1, t) = 0$.
The initial condition is given by:
\begin{equation}
    u(x, 0; \theta) = - \sin(\pi x)
    \label{eq:parametric_ic_noisy}
\end{equation}
We have the neural network find a PINN model and then employ the GAM to learn the residuals of the PINN loss. This improves the optimization process, allowing us to correctly re-discover Burgers' equation. We visually show the effect of adding different noise levels with the corresponding models, demonstrating how GAM can find the correct equation.

While the neural network's training is identical to the previous experiments, random noise is added to all the data points. Hyper-parameter $p$ controls how much the PDE is jittered, e.g., 0.2*[-1, 1].
The residuals for the GAM in this case are defined as $r = E(\hat{u}) - E(\tilde{u})$, see (Eq.~4), where $E(\tilde{u})$ represents the noisy PDE. The GAM's role is to recover the original PDE.
Figure 5 shows the equation's solution with random noise at the top and the corrected de-noised solution at the bottom.
The color bar on the right shows velocity $u\in[-1,1]$ values for the density plot; the plot describes how the velocity changes in time and space.
For reference, Figure~2 shows the ground truth. In Figure~5, we can observe how adding a small amount of noise ($5\%$) causes the equation to diverge for points in the ranges $0 < x < 0.15$ and $t < 0.5$, while this is not the case with GAM. Figure~6 shows a higher level of noise ($20\%$);  the solution range is thinner [-0.7, 0.7] compared to the original [-1, 1]. We can also see that the solution $u$ is incorrectly diverging in the ranges $-0.25 < x < 0.25$ and $t < 0.6$ while the GAM model corrects for the noise effect. Even though the GAM can fix the jittering, the solution is not exact (compared to Figure~2). 

\vspace{-8mm}
\begin{figure}[ht!]
\centering
\begin{subfigure}{\linewidth}
    \centering
    \includegraphics[width=0.8\columnwidth]{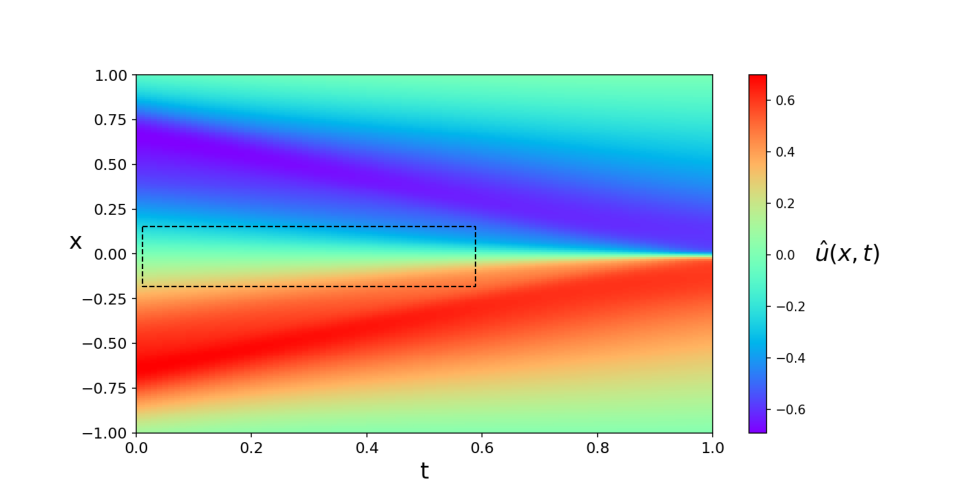}
    \label{fig:subfigAa}
\end{subfigure}
\vskip -0.1in
\begin{subfigure}{\linewidth}
    \centering
    \includegraphics[width=0.8\columnwidth]{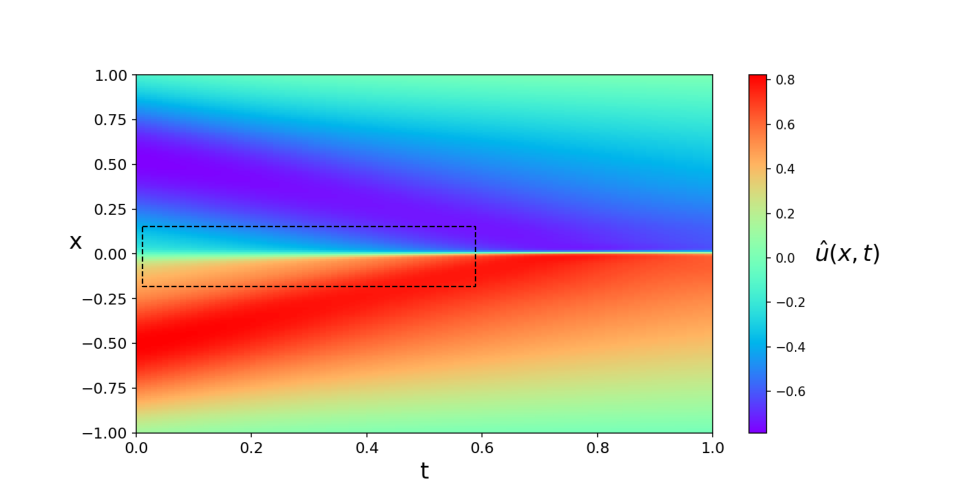}
    \label{fig:subfigBb}
\end{subfigure}
\vspace*{-4mm}
\caption{Burgers equation: comparison of the noisy solution u(x,t) (top) with de-noised solution (bottom) for initial conditions (Eq.~15) with noise p = 20\%.}
\label{fig:subfigures2}
\end{figure}

\vspace*{-4mm}

\section{Conclusions and Future Work}

This paper proposes a new method for meta-learning loss functions of parametrized PDEs by incorporating an additive regression model that minimizes the residuals. The learned loss is a regularization term for the global model that smooths out the residuals.

The experimental results of this paper suggest that learning a loss function for meta-learning PDEs improves the convergence and performance of the meta-learner. Specifically, by testing Burger's and the 2D heat equation, we observe how our approach, \GAM, outperforms \MAML\ and Random Weighting when testing on new parametric initial conditions. When we use initial conditions (Eq.~14) for the 2D heat equation, \GAM\ -in contrast to \MAML- outperforms Random Weighting. Finally, we show how \GAM\ can be used to de-noise a PDE. We point to significant gains obtained when learning loss functions for solving parametric PDEs using PINNs.

We recognize that GAMs have limitations and cannot properly discover complex analytical equations. In future work, we will develop techniques for discovering missing parts of a PDE or loss functions \cite{XPINN}, \cite{PDE-LEARN}, e.g., advanced symbolic regression techniques using neural networks. We plan to apply our results to discover analytical partial differential equations from experimental data across scientific domains.

\begin{credits}
\subsubsection{\discintname}
The authors have no competing interests to declare that they are relevant to the content of this article.
\end{credits}
%
%
%
\bibliography{bibliography}
\end{document}